\documentclass[letterpaper, 10 pt, conference]{ieeeconf}  

\IEEEoverridecommandlockouts                              

\usepackage{graphicx} 
\usepackage{amsmath} 
\usepackage{amssymb}  
\usepackage{booktabs}
\usepackage{cite}
\usepackage{subcaption}
\usepackage{comment}
\usepackage{url}

\title{\LARGE \bf
Increasing the Efficiency of DETR for Maritime High-Resolution Images
}
\author{Tinsae~Yehuala$^{1}$, Cheng~Hao$^{1}$,~\textit{Member,~IEEE},
  and Ville~Lehtola$^{1}$,~\textit{Member,~IEEE}%
  \thanks{$^{1}$Dept.\ of Earth Observation Science, ITC Faculty,
    University of Twente, Enschede, The Netherlands.
    {\tt v.v.lehtola@utwente.nl}}%
}
\begin{document}

\maketitle
\thispagestyle{empty}
\pagestyle{empty}


\begin{abstract}

Maritime object detection is critical for the safe navigation of unmanned surface vessels (USVs), requiring accurate recognition of obstacles from small buoys to large vessels. 
Real-time detection is challenging due to long distances, small object sizes, large-scale variations, edge computing limitations, and the high memory demands of high-resolution imagery. 
Existing solutions, such as downsampling or image splitting, often reduce accuracy or require additional processing, while memory-efficient models typically handle only limited resolutions. 
To overcome these limitations, we leverage Vision Mamba (ViM) backbones, which build on State Space Models (SSMs)  to capture long-range dependencies while scaling linearly with sequence length.
Images are tokenized into sequences for efficient high-resolution processing. 
For further computational efficiency, we design a tailored Feature Pyramid Network with successive downsampling and SSM layers, as well as token pruning to reduce unnecessary computation on background regions. 
Compared to state-of-the-art methods like RT-DETR with ResNet50 backbone, our approach achieves a better balance between performance and computational efficiency in maritime object detection.

\end{abstract}


\section{INTRODUCTION}
Maritime object detection is crucial for enabling unmanned surface vessels (USVs) to navigate safely by accurately recognizing and localizing potential obstacles.
However, deploying deep learning-based object detection methods \cite{rcnn_svm,ren_faster_2016,he_mask_2018,dosovitskiy_image_2021,redmon_yolo9000_2016,zhao_detrs_2024} in real time presents several significant challenges.
For instance, vessels up to 50\,meters wide are expected to be detected from as far as 8 nautical miles (NM), while smaller targets such as port buoys must be detected at 1\,NM \cite{ville}.
The long distances, small object sizes, and large-scale variations make maritime object detection particularly challenging.
Although high-resolution imagery \cite{singh_effect_2024, wang_megadetectnet_2023} provides fine-grained details that improve detection accuracy, it also requires substantial memory and leads to longer inference times for deep learning models.
These limitations are especially problematic for USVs, which often depend on edge devices with limited computational resources to perform real-time processing.

\begin{figure}[t!]
    \centering
    \includegraphics[width=1\linewidth]{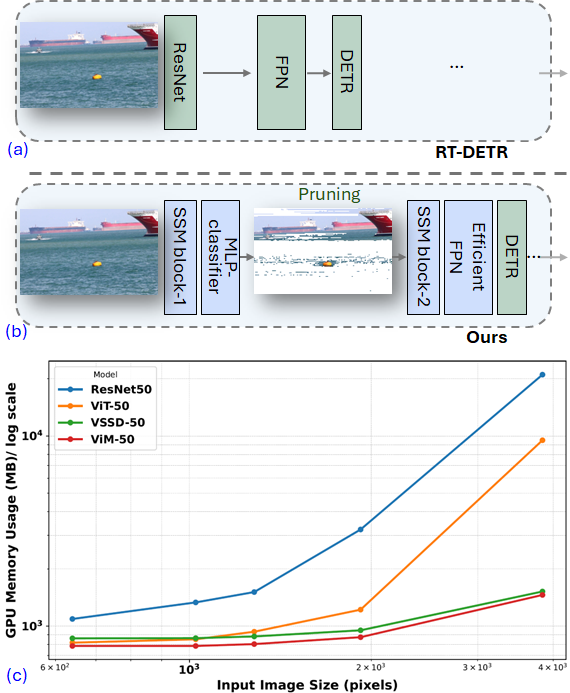}
    \caption{To achieve more efficient maritime object detection with high-resolution images compared to the current state-of-the-art RT-DETR \cite{zhao_detrs_2024} based detector (a), our method leverages ViM \cite{zhu_vision_2024} with state space models (SSMs) \cite{mamba_seq1, mamba_seq2} and a tailored Efficient FPN to emphasize the foreground containing target objects by pruning the background tokens with an MLP classifier (b). By incorporating State Space-based vision models, such as VSSD-50 \cite{vssd} and ViM-50 \cite{zhao_detrs_2024}, our approach achieves more efficient GPU usage as the input image size increases (c).}
    \label{fig:teaser}
    \vspace{-16pt}
\end{figure}

The high-resolution images pose a unique challenge. While the computational capabilities of model components have improved, these models are typically trained and benchmarked at relatively low resolutions, for example, $640\times640$ for YOLO-based \cite{redmon_you_2016, redmon_yolo9000_2016} detectors and $800\times1333$ for R-CNN-based \cite{ren_faster_2016, he_mask_2018} detectors. However, maritime images often have much higher resolutions, e.g., $1920\times1080$ (=2Mpixel) \cite{singaporez}, and CNN-based backbones scale quadratically in memory and computation with image size, making inference at such resolutions substantially more expensive and often infeasible on edge devices with limited GPU memory. From expert consultations, the commercial USVs are preferred to be equipped with multiple cameras with even larger resolution than 2Mpixel, up to 6 or 10 Mpixel, for 360-degree situational awareness.

Specific strategies have been proposed to deal with high-resolution images.
One approach to addressing the high computational resource requirement is to use a downsampled version of high-resolution images.
However, small objects inherently contain limited distinguishing features.
Downsampling makes detecting small objects challenging, as it can cause them to become indistinguishable from background elements or even disappear entirely \cite{li_saccadedet_2024, wang_megadetectnet_2023}.
Another possible solution is to divide a high-resolution image into smaller sub-images and perform detection.
If this requires additional effort to stitch sub-images for large objects spanning multiple regions, it is not optimal for large maritime object detection. However, if the image is tokenized into a sequence while keeping the positional encodings, it is an attractive approach for small to large object detection.

Recently, Vision Mamba (ViM) \cite{zhu_vision_2024} was introduced as a vision backbone model built on State Space Models (SSMs) \cite{mamba_seq1, mamba_seq2}, extending Mamba’s linear-scaling sequence modeling to capture long-range dependencies in visual data.
In this paper, we leverage the linear scaling properties of ViM for efficient maritime object detection, while tokenizing the high-resolution images into long sequences.
Images are transformed into \(R^{T\times D}\) form, where \(D\) is the embedding dimension and $T$ is the sequence length.
Each patch is linearly projected into a (vector) token of dimension \(D\) using a shared matrix weight.
Compared to CNN-based methods \cite{rcnn_svm, ren_faster_2016, he_mask_2018}, and Transformer-based methods \cite{liu_swin_2021, zhu_deformable_2021, zhao_detrs_2024}, this tokenization enables a more efficient way of handling high-resolution images, as the SSM-based processing of the resulting sequence scales linearly, and it enables additional modifications.

We design a tailored Feature Pyramid Network (FPN) \cite{lin_feature_2017} to handle the large-scale variations of maritime objects, observe the size differences of the ships and the buoy in Fig. \ref{fig:teaser}b. The Mamba tokenization scheme enables our efficient feature pyramid to avoid upsampling, in contrast to the original FPN (Fig. \ref{fig:teaser}a), by applying successive downsampling with depthwise convolutions and introducing SSM layers between the downsampling stages. 

Furthermore, in maritime environments, objects are sparsely and unevenly distributed, and large portions of the image consist of sky and water, resulting in wasted computational effort on tokens containing mostly background regions.
To address this, we employ pruning of the Mamba tokens to reduce computation on less important background areas in marine images (Fig. \ref{fig:teaser}b).
Compared to the state-of-the-art results on transformer-based model RT-DETR with a ResNet50 backbone \cite{zhao_detrs_2024}, our approach achieves more efficient GPU usage as the input image size increases (Fig. \ref{fig:teaser}c).

The main contributions of this work are:
\begin{itemize}
    \item We propose a Mamba-based framework that leverages an efficient Mamba backbone and integrates a Selective State Space model for maritime object detection. Unlike downsampling approaches, our method preserves the original high-resolution images, which is beneficial for fine-grained information for object detection. 
    \item Detection is guided toward foreground regions through background token pruning, reducing computational costs (and thereby energy) by skipping large empty areas of sky and water that dominate maritime imagery.
    \item We design an efficient feature pyramid network that combines Structured State Space models (SSMs) with depthwise convolutions, eliminating the need for inefficient upsampling while effectively capturing fine-grained details crucial for small object detection.  
    \item Our approach achieves a favorable trade-off between accuracy and efficiency. Compared to the state-of-the-art model, it delivers a $6\times$ speedup in inference and a 50\% reduction in memory usage on a 4GB NVIDIA Quadro T600 Mobile GPU, with only a minor performance drop in small object detection.      
\end{itemize}

\section{Related work}

\noindent\textbf{Maritime Object Detection}.
As one of the key applications of generic object detection \cite{er2023ship}, deep learning-based object detectors such as R-CNN-based \cite{rcnn_svm, ren_faster_2016, he_mask_2018} two-stage, YOLO-based \cite{redmon_you_2016, redmon_yolo9000_2016} one-stage, and Transformer-based \cite{liu_swin_2021, zhu_deformable_2021, zhao_detrs_2024} models have been widely explored for maritime object detection. Representative models include YOLO-MRS \cite{yu2024yolo}, which combines YOLOv8 \cite{varghese2024yolov8} with multi-scale cross-axis attention to balance accuracy with computational efficiency for real-time deployment; YOLOv5s \cite{liu2024yolov5s}, which introduces a lightweight YOLOv5 with Swin Transformer \cite{liu_swin_2021} to enhance feature extraction under challenging maritime conditions; and HMPNet \cite{zhang2025hmpnet}, which presents a hierarchical multi-path feature aggregation architecture for shipborne object detection, focusing on capturing multi-scale features and improving detection from the perspective of moving platforms. Despite these advancements, a notable research gap remains: most existing models prioritize either accuracy or efficiency but struggle to process high-resolution maritime imagery in real time, which is critical for detecting small or distant targets without compromising computational feasibility.


Moreover, in the maritime domain, there is no universally accepted benchmark dataset, although several datasets exist \cite{su2023survey, jungbauer2025maritime}. As a result, many studies \cite{custom, custom2} rely on proprietary or custom datasets that are not publicly available, limiting reproducibility and comparison across methods. 
Examples of open datasets include ABOship \cite{abo} containing 9,880 images \(1920\times720\) depicting inland water scenes, and the Singapore Maritime Dataset \cite{singaporez} including 4,445 higher-resolution images (\(1920\times1080\)) captured in open sea environments. 
In this work, we focus on open sea USV deployment and select SMD accordingly; its high resolution and scene characteristics directly motivate our background pruning approach.

\begin{figure*}[t!]
    \centering
    \includegraphics[width=1\linewidth]{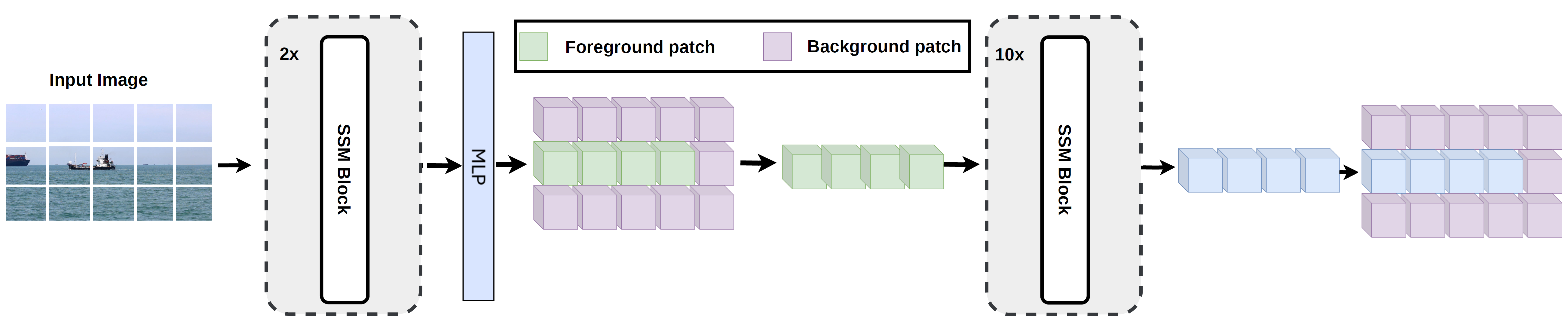}
    \caption{Background-Aware Linear-Scaling Backbone. The flattened image is first passed through a shallow pre-backbone composed of SSM blocks. A shallow MLP then classifies the patches into foreground and background tokens. The foreground tokens are forwarded to the main backbone, also built with SSM blocks, which is responsible for extracting the feature map. To avoid extra computation, the background tokens are not processed further but retain their pre-backbone representations.}
    \label{fig:backbone}
\end{figure*}

\noindent\textbf{Object Detection with High-Resolution Images}.
Traditional object detection methods, whether based on CNNs or transformers, face challenges when dealing with high-resolution inputs due to excessive GPU memory requirements and slow inference speeds. To address this, recent works have adopted coarse-to-fine strategies, where the model first predicts regions with a high probability of containing objects, then crops those regions for fine-grained classification and bounding box regression at lower resolutions \cite{li_saccadedet_2024, wang_megadetectnet_2023}. While effective, these approaches require initial downsampling, which negatively impacts the detection of small or distant objects.
Other methods, such as \cite{liu_esod_2024}, attempt to process high-resolution inputs directly without downsampling, but still rely on dense backbones that are computationally expensive. 
However, maritime object detection requires high-resolution images for detecting objects ranging from small buoys to large cargo ships with limited computational resources \cite{wang_megadetectnet_2023,li_saccadedet_2024,singaporez}. Therefore, in this paper, we leverage a Mamba-based backbone \cite{mamba_seq1, mamba_seq2, zhu_vision_2024} that tokenizes high-resolution images into long sequences without trading off the requirements of resolution and efficiency.

\vspace{4pt}
\noindent\textbf{State Space-based Vision Models}.
Recent advancements in real-time object detection have focused on improving inference efficiency while reducing memory usage. CNN-based detectors, such as YOLO and SSD \cite{liu_ssd_2016, redmon_you_2016}, achieved early success, but their dense operations limit scalability to high-resolution inputs. Transformer-based detectors, such as DETR \cite{carion_end--end_2020} and its variants \cite{zhao_detrs_2024, zong_detrs_2023}, alleviate some issues by adopting anchor-free and non-NMS training strategies, enabling sparse predictions and improved computational efficiency. However, transformers still suffer from quadratic complexity with respect to input tokens.
To address this, various strategies have been proposed. Shifted window transformers \cite{liu_swin_2021} and deformable transformers \cite{zhu_deformable_2021} reduce quadratic costs by restricting attention to local regions or deformable receptive fields. Token pruning \cite{roh_sparse_2022, zheng_less_2023, sah2024token} further decreases computational burden by discarding uninformative tokens during inference, though many approaches still rely on CNN or transformer backbones.
A promising alternative is the adoption of state space models (SSMs) for vision tasks. Vision Mamba (ViM) \cite{zhu_vision_2024} introduces a backbone that captures long-range dependencies while scaling linearly with input size, making it particularly suited for high-resolution detection.
VSSD \cite{vssd} extends ViM by replacing SSM blocks with a non-causal state space duality formulation, preserving linear scaling while improving accuracy.
In this work, we incorporate a background-aware linear-scaling backbone with token pruning, aiming to achieve real-time efficiency without the quadratic complexity of transformers.

\section{Method}
In this section, we first briefly revisit the framework of Vision Mamba (ViM) \cite{zhu_vision_2024}, which handles a high-resolution image input as a sequence of image patches in Sec. \ref{sec:vim}.
Then, we introduce our pipeline of background pruning to guide the network on foreground tokens in Sec. \ref{sec:mamba-backbone} and our efficient Feature Pyramid module for maritime object detection in Sec. \ref{FPN}. 
In the end, we introduce the loss functions in Sec. \ref{sec:losses}.

\subsection{\textbf{Preliminary of Vision Mamba}} \label{sec:vim}

\begin{figure}
    \centering
    \includegraphics[width=1.\linewidth]{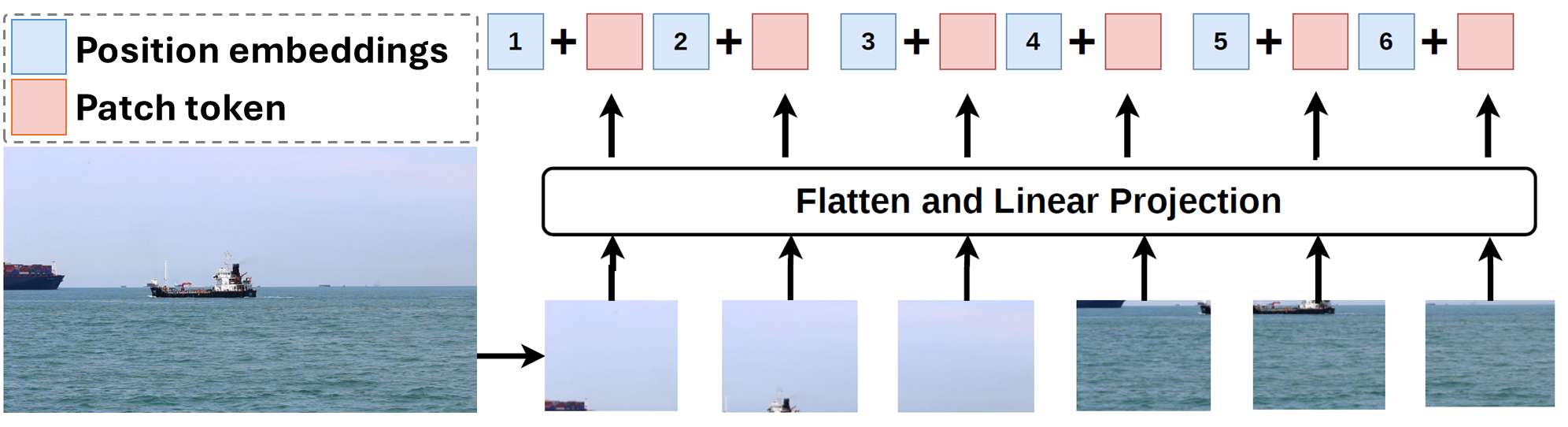}
    \caption{Tokenization. The input image is divided into patches, and a shared matrix weight is used to project each patch into a vector, creating a flattened sequence of vectors. A positional encoding vector is added to encode the spatial position of each token in the original image.}
    \label{tokenization}
\end{figure}

Vision Mamba (ViM) \cite{zhu_vision_2024} is a vision backbone built on state space models (SSMs) \cite{mamba_seq1, mamba_seq2}, offering an efficient alternative to self-attention. 
As shown in Fig. \ref{tokenization}, in ViM, an image is tokenized into patch embeddings \(u_t = \operatorname{Embed}(X_t) + p_t\), where $t$ is the position of the input and \(p_t\) denotes the position encoding. 
An SSM in discrete recurrence form is formulated as
\begin{align}
x_t &= \bar{A} x_{t-1} + \bar{B} u_t, &
y_t &= \bar{C} x_t + \bar{D} u_t,
\end{align}
where \(\bar{A},\bar{B},\bar{C},\bar{D}\) are discrete matrices.

To capture spatial context information of the input image, two SSM streams are run in both forward ($f$) and backward ($b$) directions:
\begin{align}
x_t^{(f)} &= \bar{A}^{(f)} x_{t-1}^{(f)} + \bar{B}^{(f)} u_t, \\
y_t^{(f)} &= \bar{C}^{(f)} x_t^{(f)} + \bar{D}^{(f)} u_t, \\
x_t^{(b)} &= \bar{A}^{(b)} x_{t+1}^{(b)} + \bar{B}^{(b)} u_t, \\
y_t^{(b)} &= \bar{C}^{(b)} x_t^{(b)} + \bar{D}^{(b)} u_t.
\end{align}

The outputs are fused to form the bidirectional representation:
\begin{align}
h_t = \phi\big(W_f y_t^{(f)} + W_b y_t^{(b)}\big),
\end{align}
where \(W_f, W_b\) are learnable fusion parameters and \(\phi\) is a nonlinear activation (e.g., GELU). Stacking such blocks yields global visual representations at lower cost than self-attention, scaling as \(\mathcal{O}(TD)\) in time and \(\mathcal{O}(D)\) in memory, compared to \(\mathcal{O}(T^2)\) for transformers (both time and memory). $D$ denotes the state dimension of the SSM, and $T$ is the sequence length. This tokenization into a sequence, processed by SSMs with linear complexity, is particularly suited for handling high-resolution maritime images.

\subsection{\textbf{Background Pruning}} \label{sec:mamba-backbone}

We propose a pruning module to reduce unnecessary computation on background tokens. 
As shown in Fig. \ref{fig:backbone}, the input image is split into patches and flattened into tokens \(R^{T\times D}\).
Each token is then passed through a stack of SSM layers (2 layers). The output is then passed through a shallow multilayer perceptron (MLP) classifier to produce a confidence score that indicates whether the corresponding patch is a foreground patch or a background patch. Patches that score the top \(K\) confidence scores are considered foreground. Specifically, $K=(1-r)T$, where $r$ is the pruning ratio (a fixed hyperparameter; $r=0.5$ in our main experiments, see Table \ref{tab:ablation_networkdesign} in Results). The SSM processing is selective. The foreground patches are passed to another stack of SSM blocks (10 layers), which further extracts the features. The background tokens are not processed further but retain their pre-backbone representations and are concatenated back at their original sequence positions before being passed to the FPN.


To distinguish between background and foreground patches, labels for the patches are required for the supervised training. However, maritime datasets do not come with ground truth labels for each patch. 
We follow \cite{zhu_vision_2024, zheng_less_2023} to use the ground truth bounding boxes to create training labels as follows:
\begin{equation}
L_i =
\begin{cases}
1, & \text{if } J(P_n, B_m)  \neq 0,  \\
0, & \text{if } J(P_n, B_m) = 0.
\end{cases}
\end{equation}

\begin{equation}
J(P_n, B_m) = \frac{|P_n \cap B_m|}{|P_n \cup B_m|},
\end{equation}
where \(P_n\) is the \(n^\text{th}\) patch and \(B_m\) is the \(m^\text{th}\) bounding box, \(L_n\) is the assigned label for \(n^\text{th}\) token, \(J(P_n , B_m )\) denotes intersection over union.

\subsection{\textbf{Efficient Feature Pyramid Module}} \label{FPN}
\begin{figure}
\centering
\begin{subfigure}{.5\linewidth}
  \centering
  \includegraphics[width=\linewidth]{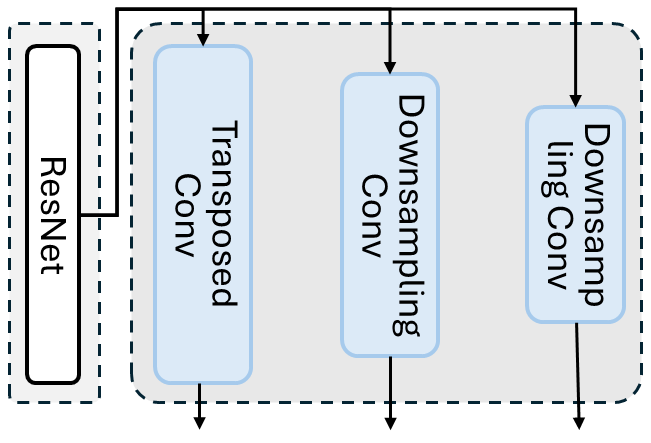}
  \caption{FPN used in RT-DETR \cite{zhao_detrs_2024}}
  \label{fig:simple-FPN}
\end{subfigure}%
\begin{subfigure}{.5\linewidth}
  \centering
  \includegraphics[width=\linewidth]{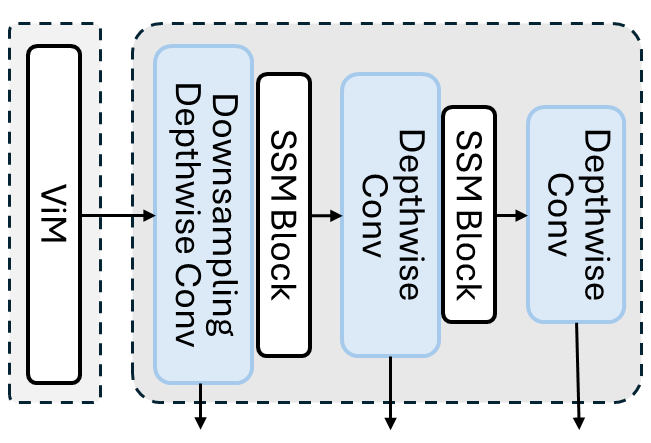}
  \caption{Our efficient FPN}
  \label{fig:efficient-FPN}
\end{subfigure}
\caption{Comparison between the FPN used in RT-DETR and our proposed efficient FPN. In RT-DETR, the FPN takes the last feature map from the backbone and processes it in parallel through upsampling and downsampling. In contrast, our efficient FPN avoids upsampling by applying successive downsampling using depthwise convolutions and introducing SSM layers between the downsampling stages.}
\label{fig:test}
\end{figure}

In order to capture the features of objects at different resolutions across multiple scales in size, we follow RT-DETR to pass the output of the SSM layers to a Feature Pyramidal Network (FPN) \cite{simple_fpn}.
However, unlike RT-DETR, which employs the standard FPN (Fig. \ref{fig:simple-FPN}) that includes time-consuming upsampling in the top-down pass, we design an efficient FPN (Fig. \ref{fig:efficient-FPN}) by further leveraging the efficiency of SSM layers. Namely, we replace the transposed convolutional layer with a depthwise convolutional layer coupled with an SSM block. Our Efficient-FPN processes features sequentially, omitting upsampling. 
Although upsampling is omitted, the SSM layers inserted between downsampling stages maintain a global receptive field over the full token sequence, enabling high-level semantic information to propagate across scales without an explicit upsampling path.
This design further increases memory efficiency and inference speed.

The extracted features from the efficient FPN are fed into a DETR-based detector, similar to the RT-DETR method \cite{zhao_detrs_2024}.
Compared to the original RT-DETR with its ResNet backbone, our pipeline achieves higher computational efficiency by discarding background patches for object detection and eliminating the costly convolutional layers.

\subsection{\textbf{Loss Functions}} \label{sec:losses}

\noindent\textbf{Detection Loss.} Following \cite{zhao_detrs_2024}, the bipartite algorithm is used for matching the prediction to the ground truth as follows:
\begin{equation}
\hat{\sigma} = \arg\min_{\sigma \in S_N} \sum_{i=1}^{N} \mathcal{L}_{\text{match}}(y_i, \hat{y}_{\sigma(i)}),
\end{equation}
where \(y\) is the ground truth and \(\hat{y}\) is a prediction, \(S_N\) is a set that contains all the permutations of the matching between \(y\) and \(\hat{y}\), \(\sigma\) is a possible matching in the \(S_N\), and \(\hat{\sigma}\) is the optimal match. 
\(\mathcal{L}_\text{match}(y_i, \hat{y})\) is the matching cost given by
\begin{equation}
   \mathcal{L}_\text{match}(y_i, \hat{y}_{\sigma(i)}) = -\hat{p}_{\sigma(i)}(c_i) + \mathcal{L}_\text{box} (b_i, \hat{b}_{\sigma (i)}),
\end{equation}
where \(c_i\) is target class label, \(\hat{p}_{a(i)}(c_i)\) is the probability of class \(c_i\), and \(\mathcal{L}_\text{box} (b_i, \hat{b}_{\sigma (i)})\) is the bounding box prediction loss.

After finding the match between the prediction and the ground truth, the Hungarian loss function is used as the objective function for training, defined as
\begin{equation}
    \mathcal{L}_\text{Hungarian}(y, \hat{y}) = \sum^N_{i=1} [-log\hat{p}_{\hat{\sigma}(i)}(c_i) + \mathcal{L}_\text{box}(b_i, \hat{b}_{\hat{\sigma}})(i)].
\end{equation}

\noindent\textbf{Patch Classification Loss.} We use a binary cross-entropy to train the foreground and background patch classifier. 
However, the number of background batches is much higher than that of the foreground batches, creating a class imbalance. Therefore, the binary cross-entropy is counter-weighted to prevent class imbalance as follows:
\begin{equation}
\mathcal{L}_\text{patch} = -\frac{1}{M} \sum_{j=1}^{M} 
\Big[ w \, y_j \log(\hat{p}_j) + (1 - y_j) \log(1 - \hat{p}_j) \Big],
\end{equation}
where \(\hat{p}_j\) is the estimated probability from the shallow MLP classifier, \(y\) is the ground truth label and the weight \(w\) is defined as
\begin{equation*}
w = \frac{M_{\text{back}}}{M_{\text{fore}}+\epsilon},
\end{equation*}
where \( M_\text{back}\) is the number of background patches, \(M_\text{fore}\) is the number of foreground patches. The weight $w$ is recomputed per-batch (across all flattened patches in the batch) for every forward pass, and $\epsilon=10^{-6}$ prevents divisions by zero.

\section{Experiments and Results}

\subsection{Experimental setup}

\noindent\textbf{Dataset.}
The Singapore maritime dataset \cite{singaporez}, available in COCO format from the Roboflow website \cite{roboflow}, with input size of \(1920\times1080\), is used to train the models. The training, validation, and test sets contain 4445, 1271, and 636 images, respectively. During training, no augmentation technique is applied to the images. 

\vspace{2pt}
\noindent\textbf{Evaluation Metrics.}
We aim to assess the efficiency and accuracy of maritime object detection. 
The performance metrics that quantify the performance criteria are maximum GPU memory usage, Giga Floating Point Operations (GFLOPs), frames per second (FPS), and mean average precision (mAP). 
Moreover, following the COCO framework, mAP\_s, mAP\_m, and mAP\_l are used as performance evaluations for small, medium, and large object detection to further reveal the performance of maritime objects of different sizes. Here, small, medium, and large refer to the spatial resolution of the bounding boxes, which are \(< 32\times32\), \(32\times32 - 96\times96\), and \(> 96\times96\), respectively.

\vspace{2pt}
\noindent\textbf{Comparison models.}
We benchmark the performance of our model against RT-DETR \cite{zhao_detrs_2024}, a real-time Detection Transformer-based (DETR) \cite{carion_end--end_2020} detector that has demonstrated excellent efficiency and accuracy on generic object detection across various datasets.  
We select RT-DETR as the primary baseline for two reasons. First, RT-DETR has been shown to outperform YOLO-based detectors on real-time object detection benchmarks \cite{zhao_detrs_2024}. Second, RT-DETR is released under the Apache 2.0 license, which is more suitable for commercial USV deployment than the AGPL-3.0 license of the Ultralytics YOLO family. 
Moreover, we also implement our model with Vision Transformers (ViT) \cite{dosovitskiy_image_2021} and Vision State Space Duality (VSSD) \cite{vssd}, both of which share the same patch tokenization strategy; the framework adapts to each by substituting the SSM blocks in the backbone with the respective model's computational blocks. 

\vspace{4pt}
\noindent\textbf{Training setup.}
For RT-DETR, due to the large input image size, we used a batch size of 6 to train the model with ResNet50 on the Singapore maritime dataset, while the other training hyperparameters were kept the same as the default setting \cite{zhao_detrs_2024}.
For the other models, the batch size was set to 8 due to their more efficient memory usage.
The patch size was set to \(32\times32\) to balance the accuracy and efficiency.
The learning rate was set to \(1\times10^{-4}\) with a weight decay \(1\times10^{-4}\).
All the models were trained for 100 epochs using the AdamW optimizer. All the models were tested on a 4GB NVIDIA Quadro T600 Mobile GPU. The code is available at \url{https://github.com/vlehtola/maritime_object_detection.git}.

\subsection{Results}
Our models demonstrate competitive performance compared to RT-DETR on the Singapore Maritime dataset. On the test set (Table~\ref{tab:result_test}), Our-VSSD achieves the highest overall mAP of 79.0, improving over RT-DETR (75.3), with particularly strong performance on small objects (mAP\_s = 73.1). Our-ViM achieves the best performance on large objects (mAP\_l = 93.7), while Our-ViT underperforms relative to the other models. On the validation set (Table~\ref{tab:result_validation}), Our-VSSD again achieves the highest overall mAP (79.4), whereas Our-ViM performs best on large objects (mAP\_l = 91.9). RT-DETR maintains strong performance on medium objects in both sets but is generally outperformed by our VSSD- and ViM-based models in overall accuracy. 

\begin{table}[ht]
\centering
\caption{Results on the Singapore Maritime test set. Best values are highlighted in \textbf{boldface}.}
\vspace{-4pt}
\small
\begin{tabular}{l|c|cccc}
\toprule
Model   & Backbone & mAP  & mAP\_s & mAP\_m & mAP\_l \\ \midrule
RT-DETR & ResNet50 & 75.3 & 70.9   & \textbf{78.0}   & 86.8   \\
Our-VSSD & VSSD     & \textbf{79.0} & \textbf{73.1}   & 77.5   & 85.3   \\
Our-ViT & ViT      & 70.2 & 62.8   & 67.0   & 88.4   \\
Our-ViM & ViM      & 78.4 & 71.0   & 75.1   & \textbf{93.7}   \\ \bottomrule
\end{tabular}
\label{tab:result_test}
\vspace{-6pt}
\end{table}

\begin{table}[ht]
\centering
\caption{Results on the Singapore Maritime validation set. Best values are highlighted in \textbf{boldface}.}
\vspace{-4pt}
\small
\begin{tabular}{l|c|cccc}
\toprule
Model   & Backbone & mAP  & mAP\_s & mAP\_m & mAP\_l \\ \midrule
RT-DETR & ResNet50 & 73.3 & \textbf{74.8}   & \textbf{74.5}   & 84.5   \\
Our-VSSD & VSSD     & \textbf{79.4} & 70.5   & 72.5   & 89.6   \\
Our-ViT & ViT      & 67.6 & 61.7   & 62.2   & 85.5   \\
Our-ViM & ViM      & 76.0 & 68.3   & 73.2   & \textbf{91.9}   \\ \bottomrule
\end{tabular}
\label{tab:result_validation}
\vspace{-6pt}
\end{table}

\begin{table}[hbpt]
\centering
\caption{Computational efficiency measured in number of parameters, GFLOP, and FPS. Best values are highlighted in \textbf{boldface}.}
\vspace{-4pt}
\small
\begin{tabular}{l|c|ccc}
\toprule
Model   & Backbone & \#Param. & GFLOP & FPS \\ \midrule
RT-DETR & ResNet50 & 42.7M    & 606   & 1.1 \\
Our-VSSD & VSSD     & 45.6M    & 45    & 6.1 \\
Our-ViT & ViT      & 42.7M    & 52    & 3.6 \\
Our-ViM & ViM      & \textbf{41.0M}    & \textbf{40}    & \textbf{6.9} \\ \bottomrule
\end{tabular}
\label{tab:result_efficiency}
\end{table}

\begin{table*}[t!]
\centering
\caption{Ablation study on the network design using the ViM backbone.}
\small
\begin{tabular}{l|ccc|cccc|cc}
\toprule
No. & FPN                & Patch size                & Pruning       & mAP           & mAP\_s        & mAP\_m        & mAP\_l        & \#Param.       & GFLOP       \\ \midrule
1   & Simple             & \(16\times16\)           & 0\%           & 74.8          & 74.0          & 72.4          & 86.3          & 79.5M          & 784         \\
2   & Simple             & \(24\times24\)           & 0\%           & 76.2          & 71.2          & 73.8          & 90.3          & 79.5M          & 350         \\
3   & Simple             & \(32\times32\)           & 0\%           & 77.6          & 73.0          & 73.1          & 91.2          & 79.5M          & 199         \\
4   & Efficient          & \(16\times16\)           & 0\%           & 76.7          & 73.0          & 73.1          & 89.6          & 41.0M          & 215         \\
5   & Efficient          & \(16\times16\)           & 50\%          & 77.2          & 72.3          & 72.1          & 90.0          & 41.0M          & 188         \\
6   & Efficient          & \(32\times32\)           & 50\%          & 76.0          & 68.3          & 73.2          & 91.9          & 41.0M          & 40           \\ \bottomrule
\end{tabular}
\label{tab:ablation_networkdesign}
\vspace{-6pt}
\end{table*}

\begin{figure*}[t!]
    \centering
    \includegraphics[width=\linewidth]{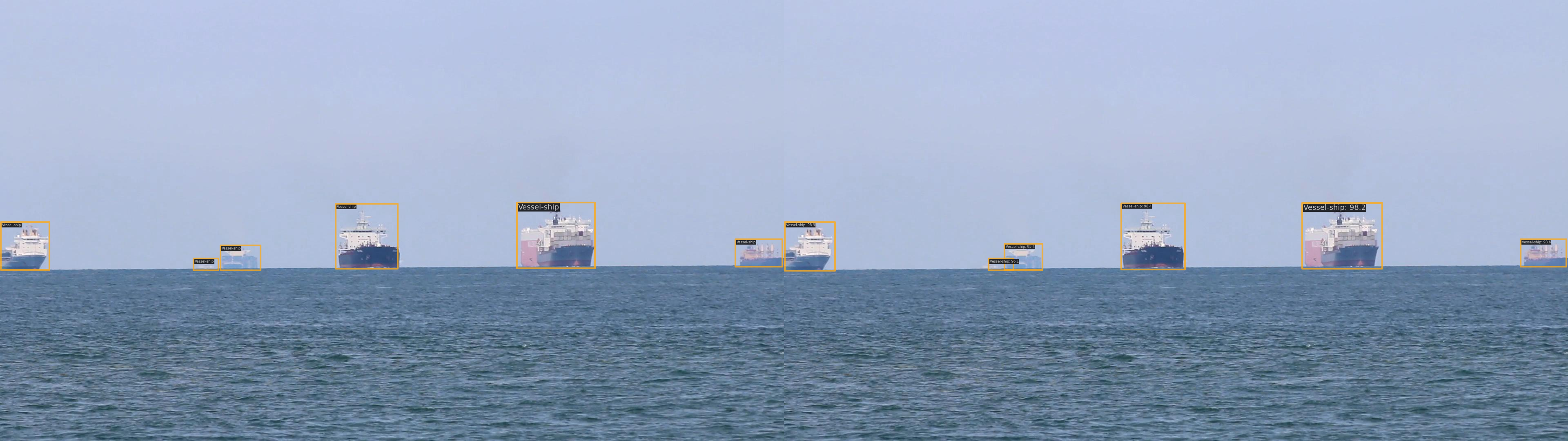}\\[0.3ex]
    \includegraphics[width=\linewidth]{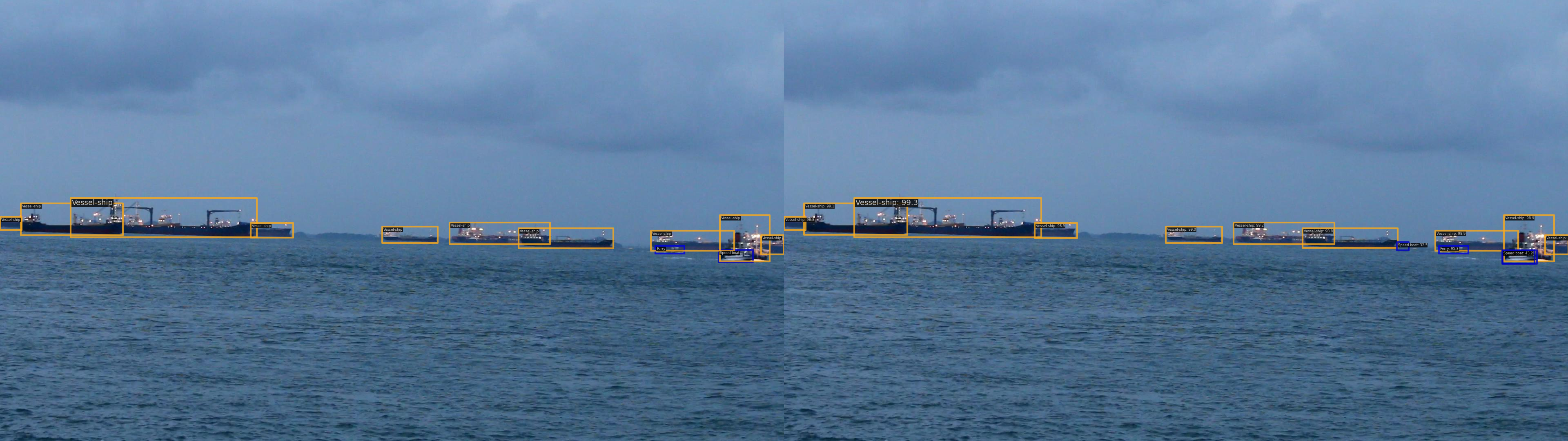}\\[0.3ex]
    \includegraphics[width=\linewidth]{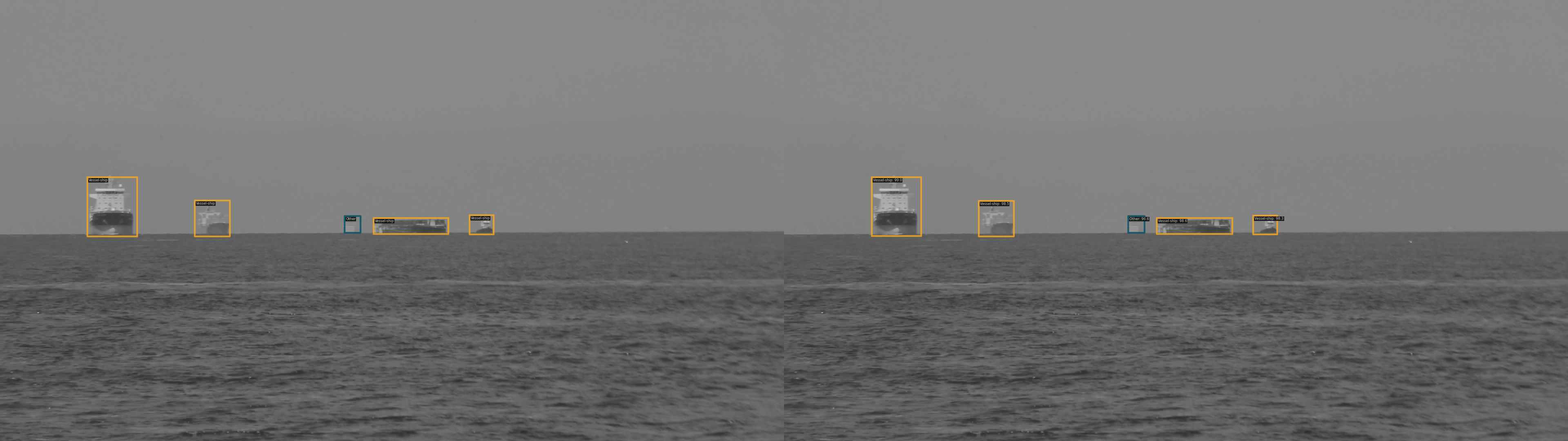}\\[0.3ex]
    \includegraphics[width=\linewidth]{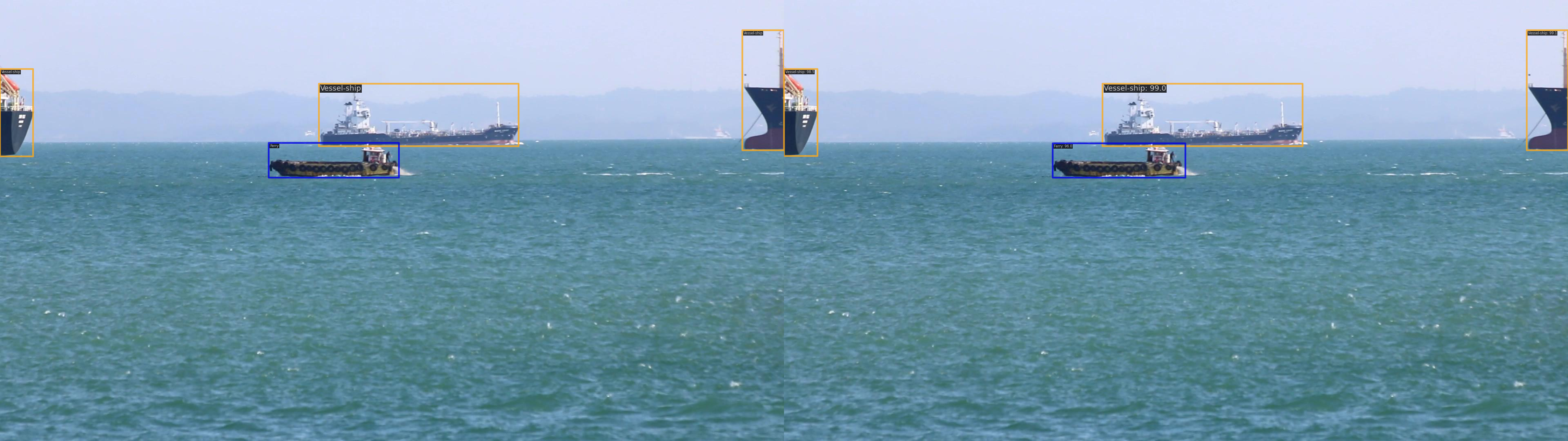}
    \vspace{-10pt}
    \caption{Visualization of maritime object detection. Left: Ground truth. Right: Prediction with the proposed method using a \(32\times32\) patch size.}
    \label{fig:appendix_qualitative}
    \vspace{-14pt}
\end{figure*}

Fig. \ref{fig:appendix_qualitative} shows the visual results against the ground truth labels. The visualization shows that maritime objects are detected, classified, and located accurately in different weather and lighting conditions. Both remote and small vessels are also detected accurately, even though some of the vessels are cluttered together from a distant perspective (second row).

In terms of computational efficiency (Table~\ref{tab:result_efficiency}), our models demonstrate significant improvements over RT-DETR. While RT-DETR with a ResNet50 backbone requires 42.7M parameters, 606 GFLOPs, and achieves only 1.1 FPS, Our-VSSD drastically reduces the computational cost to 45 GFLOPs and increases the inference speed to 6.1 FPS, despite a slightly larger parameter count (45.6M). Our-ViT also improves efficiency with 52 GFLOPs and 3.6 FPS, while Our-ViM is the most efficient, with only 41.0M parameters, 40 GFLOPs, and a high inference speed of 6.9 FPS. Overall, VSSD- and ViM-based models achieve substantial speed-ups and lower computational demands compared to RT-DETR, making them more suitable for real-time maritime object detection.

Furthermore, we inspect the GPU memory usage of the models across varying input image sizes, as shown in Fig.~\ref{fig:teaser}. 
The advantage of Mamba’s linear scaling with respect to the token sequence length 
$T=(H/Z)^2$ becomes evident at the tested resolution of $1920\times1080$. For such and higher resolutions, we recommend backbones with linear-in-$T$ scaling, since the theoretical space complexities differ significantly across architectures: transformers \(\mathcal{O}( T^2 + T D) = \mathcal{O}((\frac{H}{Z})^4 + (\frac{H}{Z})^2D)\), Mambas \(\mathcal{O}(T D + D K + D^2) = \mathcal{O}(D(\frac{H}{Z})^2 + DK + D^2)\), and CNNs \(\mathcal{O}(H^2D + K^2D^2)\). Here, the input size is \(H\times H\), the grid (patch) size is \(Z\times Z\), \(K\) is the convolution kernel size, and $D$ is the embedding dimension. CNNs grow as $\mathcal{O}(H^2)$ in memory, since they operate directly on the full-resolution pixel grid. Mambas also scale quadratically in $H$, but with respect to the token sequence length $T$, so the effective growth is divided by $Z^2$ (e.g., $1024$ for a patch size of $32$). This constant-factor reduction makes Mambas far more efficient than ResNet50 at high resolutions.

Overall, our ViM model achieves an excellent balance between detection performance and computational efficiency. It demonstrates high detection accuracy across all object sizes and superior foreground classification accuracy (Acc\_all = 85.4\%). At the same time, ViM maintains the lowest parameter count (41.0M), the lowest computational cost (40 GFLOPs), and the highest inference speed (6.9 FPS) among all evaluated models. This combination of high accuracy and efficiency makes ViM particularly well-suited for real-time maritime object detection, effectively handling both small and large objects while reducing computational overhead.

\begin{table}[hbpt]
\centering
\caption{Foreground classification accuracy. Note that most foreground patches do not contain objects.}
\vspace{-4pt}
\small
\begin{tabular}{l|cccc} 
\toprule
Model  & Acc\_all & Acc\_s & Acc\_m & Acc\_l \\ \midrule
VSSD   & 63.4     & 52.3   & 70.2   & 88.2   \\
ViT    & 34.7     & 85.1   & 89.1   & 92.9   \\
ViM    & 85.4     & 90.5   & 94.4   & 95.6   \\ \bottomrule
\end{tabular}
\label{tab:foregroundaccu}
\vspace{-10pt}
\end{table}

We performed further ablation studies on the position of the foreground classifier, the pruning ratio, and the number of SSM layers, which all yielded expected results.

\subsection{Ablation Studies}
We conduct a series of ablation studies to evaluate the impact of the efficient FPN, patch size, and pruning on detection performance and computational efficiency (Table~\ref{tab:ablation_networkdesign}).  
Replacing the simple FPN with the Efficient FPN (comparing rows 1 and 4) significantly reduces both the parameter count (from 79.5M to 41.0M) and computational cost (GFLOPs from 784 to 215) while improving overall mAP (74.8\% $\rightarrow$ 76.7\%). Varying the patch size demonstrates that larger patches reduce GFLOPs substantially (row 3 vs. row 1: 199 vs. 784 GFLOPs) with minimal impact on accuracy. Pruning further improves efficiency (row 5 vs. row 4: 188 vs. 215 GFLOPs) while maintaining competitive mAP. Overall, the combination of Efficient FPN, appropriate patch size, and pruning (row 6) achieves a strong balance between detection performance and computational efficiency, reducing GFLOPs to 40 while preserving high mAP, particularly for large objects (mAP\_l = 91.9\%). 

Furthermore, Table~\ref{tab:foregroundaccu} presents the foreground classification accuracy for different backbone models. Our-ViM achieves the highest overall accuracy (85.4\%), outperforming VSSD (63.4\%) and ViT (34.7\%). It also demonstrates consistently strong performance across all object sizes, with Acc\_s = 90.5\%, Acc\_m = 94.4\%, and Acc\_l = 95.6\%. In comparison, VSSD shows lower accuracy for small objects (Acc\_s = 52.3\%), while ViT, despite performing well on objects, struggles overall with non-object patches as it tends to label most patches as foreground, achieving high within-box accuracy but poor overall discrimination. These results highlight that ViM is highly effective at distinguishing foreground patches, which contributes to both its detection accuracy and efficiency.

\section{CONCLUSIONS}
Maritime object detection for unmanned surface vessels requires accurate recognition of both small and large obstacles from high-resolution imagery, which poses significant computational challenges for methods like RT-DETR with ResNet-50. We address this by leveraging Vision Mamba (ViM) backbones built on Selective State Space Models (SSMs), combined with tokenization, a tailored Feature Pyramid Network, and token-level pruning. This design scales linearly with the length of image patch sequence, preserves accuracy across object scales, and reduces unnecessary computation (and energy use) on the sky and water regions that dominate maritime imagery. Our proposed backbone method enables the DETR-based detector to run at over 6 FPS with the image size of \( 1920\times1080\) on an edge device. Compared to RT-DETR with ResNet-50, our method achieves a superior balance of detection performance and computational efficiency, making it suitable for practical, high-resolution maritime applications.
\section*{Acknowledgment}
This work was supported by the EU under
the Horizon Europe RIA grant agreement No. 101136006
(XTREME).

\bibliographystyle{IEEEtran}
\bibliography{bib}

@inproceedings{rcnn_svm,
  title={Rich feature hierarchies for accurate object detection and semantic segmentation},
  author={Girshick, Ross and Donahue, Jeff and Darrell, Trevor and Malik, Jitendra},
  booktitle={Proceedings of the IEEE conference on computer vision and pattern recognition},
  pages={580--587},
  year={2014}
}

@inproceedings{sah2024token,
  title={Token Pruning using a Lightweight Background Aware Vision Transformer},
  author={Sah, Sudhakar and Kumar, Ravish and Rohmetra, Honnesh and Saboori, Ehsan},
  year = {2024},
  url = {http://arxiv.org/abs/2410.09324},
  booktitle={NeurIPS 2024 Workshop on Fine-Tuning in Modern Machine Learning: Principles and Scalability}
}

@article{ren_faster_2016,
  title={Faster r-cnn: Towards real-time object detection with region proposal networks},
  author={Ren, Shaoqing and He, Kaiming and Girshick, Ross and Sun, Jian},
  journal={Advances in neural information processing systems},
  volume={28},
  year={2015}
}

@inproceedings{he_mask_2018,
  title={Mask r-cnn},
  author={He, Kaiming and Gkioxari, Georgia and Doll{\'a}r, Piotr and Girshick, Ross},
  booktitle={Proceedings of the IEEE international conference on computer vision},
  pages={2961--2969},
  year={2017}
}

@inproceedings{dosovitskiy_image_2021,
title={An Image is Worth 16x16 Words: Transformers for Image Recognition at Scale},
author={Alexey Dosovitskiy and Lucas Beyer and Alexander Kolesnikov and Dirk Weissenborn and Xiaohua Zhai and Thomas Unterthiner and Mostafa Dehghani and Matthias Minderer and Georg Heigold and Sylvain Gelly and Jakob Uszkoreit and Neil Houlsby},
booktitle={International Conference on Learning Representations},
year={2021},
}

@inproceedings{redmon_yolo9000_2016,
  title={YOLO9000: better, faster, stronger},
  author={Redmon, Joseph and Farhadi, Ali},
  booktitle={Proceedings of the IEEE conference on computer vision and pattern recognition},
  pages={7263--7271},
  year={2017}
}

@inproceedings{zhao_detrs_2024,
  title={Detrs beat yolos on real-time object detection},
  author={Zhao, Yian and Lv, Wenyu and Xu, Shangliang and Wei, Jinman and Wang, Guanzhong and Dang, Qingqing and Liu, Yi and Chen, Jie},
  booktitle={Proceedings of the IEEE/CVF conference on computer vision and pattern recognition},
  pages={16965--16974},
  year={2024}
}

@article{ville,
  title={Sensors and AI techniques for situational awareness in autonomous ships: A review},
  author = {Thombre, Sarang and Zhao, Zheng and Ramm-Schmidt, Henrik and Vallet Garcia, Jose M. and Malkamaki, Tuomo and Nikolskiy, Sergey and Hammarberg, Toni and Nuortie, Hiski and H. Bhuiyan, M. Zahidul and Sarkka, Simo and Lehtola, Ville V.},
  journal={IEEE transactions on intelligent transportation systems},
  volume={23},
  number={1},
  pages={64--83},
  year={2020},
  publisher={IEEE}
}

@article{singh_effect_2024,
  title={On the effect of image resolution on semantic segmentation},
  author={Singh, Ritambhara and Jain, Abhishek and Perona, Pietro and Agarwal, Shivani and Yang, Junfeng},
  journal={arXiv preprint arXiv:2402.05398},
  year={2024}
}

@article{wang_megadetectnet_2023,
  title={MegaDetectNet: a fast object detection framework for Ultra-High-Resolution images},
  author={Wang, Jian and Zhang, Yuesong and Zhang, Fei and Li, Yazhou and Nie, Lingcong and Zhao, Jiale},
  journal={Electronics},
  volume={12},
  number={18},
  pages={3737},
  year={2023},
  publisher={MDPI}
}

@inproceedings{zhu_vision_2024,
  title={Vision Mamba: Efficient Visual Representation Learning with Bidirectional State Space Model},
  author={Zhu, Lianghui and Liao, Bencheng and Zhang, Qian and Wang, Xinlong and Liu, Wenyu and Wang, Xinggang},
  year={2024},
  booktitle={Forty-first International Conference on Machine Learning}
}

@inproceedings{mamba_seq1,
title={Mamba: Linear-Time Sequence Modeling with Selective State Spaces},
author={Albert Gu and Tri Dao},
booktitle={First Conference on Language Modeling},
year={2024},
}

@inproceedings{mamba_seq2,
  title={Transformers are SSMs: Generalized Models and Efficient Algorithms Through Structured State Space Duality},
  author={Dao, Tri and Gu, Albert},
  year={2024},
  booktitle={Forty-first International Conference on Machine Learning}
}

@inproceedings{vssd,
  title={VSSD: Vision Mamba with Non-Causal State Space Duality},
  author={Yuheng Shi and Minjing Dong and Mingjia Li and Chang Xu},
  booktitle={Proceedings of the IEEE international conference on computer vision},
  year={2025}
}

@inproceedings{redmon_you_2016,
  title={You only look once: Unified, real-time object detection},
  author={Redmon, Joseph and Divvala, Santosh and Girshick, Ross and Farhadi, Ali},
  booktitle={Proceedings of the IEEE conference on computer vision and pattern recognition},
  pages={779--788},
  year={2016}
}

@article{singaporez,
  title={Video processing from electro-optical sensors for object detection and tracking in a maritime environment: A survey},
  author={Prasad, Dilip K and Rajan, Deepu and Rachmawati, Lily and Rajabally, Eshan and Quek, Chai},
  journal={IEEE Transactions on Intelligent Transportation Systems},
  volume={18},
  number={8},
  pages={1993--2016},
  year={2017},
  publisher={IEEE}
}

@inproceedings{li_saccadedet_2024,
  title={SaccadeDet: A Novel Dual-Stage Architecture for Rapid and Accurate Detection in Gigapixel Images},
  author={Li, Wenxi and Zhang, Ruxin and Lin, Haozhe and Guo, Yuchen and Ma, Chao and Yang, Xiaokang},
  booktitle={Joint European Conference on Machine Learning and Knowledge Discovery in Databases},
  pages={392--408},
  year={2024},
  organization={Springer}
}

@inproceedings{liu_swin_2021,
  title={Swin transformer: Hierarchical vision transformer using shifted windows},
  author={Liu, Ze and Lin, Yutong and Cao, Yue and Hu, Han and Wei, Yixuan and Zhang, Zheng and Lin, Stephen and Guo, Baining},
  booktitle={Proceedings of the IEEE/CVF international conference on computer vision},
  pages={10012--10022},
  year={2021}
}

@inproceedings{zhu_deformable_2021,
title={Deformable {DETR}: Deformable Transformers for End-to-End Object Detection},
author={Xizhou Zhu and Weijie Su and Lewei Lu and Bin Li and Xiaogang Wang and Jifeng Dai},
booktitle={International Conference on Learning Representations},
year={2021},
}

@inproceedings{lin_feature_2017,
  title={Feature pyramid networks for object detection},
  author={Lin, Tsung-Yi and Doll{\'a}r, Piotr and Girshick, Ross and He, Kaiming and Hariharan, Bharath and Belongie, Serge},
  booktitle={Proceedings of the IEEE conference on computer vision and pattern recognition},
  pages={2117--2125},
  year={2017}
}

@article{er2023ship,
  title={Ship detection with deep learning: a survey},
  author={Er, Meng Joo and Zhang, Yani and Chen, Jie and Gao, Wenxiao},
  journal={Artificial Intelligence Review},
  volume={56},
  number={10},
  pages={11825--11865},
  year={2023},
  publisher={Springer}
}

@article{yu2024yolo,
  title={YOLO-MRS: An efficient deep learning-based maritime object detection method for unmanned surface vehicles},
  author={Yu, Changdong and Yin, Haoke and Rong, Chenyi and Zhao, Jiayi and Liang, Xiao and Li, Ruijie and Mo, Xinrong},
  journal={Applied Ocean Research},
  volume={153},
  pages={104240},
  year={2024},
  publisher={Elsevier}
}

@inproceedings{varghese2024yolov8,
  title={Yolov8: A novel object detection algorithm with enhanced performance and robustness},
  author={Varghese, Rejin and Sambath, M},
  booktitle={2024 International conference on advances in data engineering and intelligent computing systems (ADICS)},
  pages={1--6},
  year={2024},
  organization={IEEE}
}

@article{liu2024yolov5s,
  title={YOLOv5s maritime distress target detection method based on swin transformer},
  author={Liu, Kun and Qi, Yueshuang and Xu, Guofeng and Li, Jianglong},
  journal={IET Image Processing},
  volume={18},
  number={5},
  pages={1258--1267},
  year={2024},
  publisher={Wiley Online Library}
}

@article{zhang2025hmpnet,
  title={HMPNet: A Feature Aggregation Architecture for Maritime Object Detection from a Shipborne Perspective},
  author={Zhang, Yu and Liu, Fengyuan and Lyu, Juan and Wei, Yi and Yu, Changdong},
  journal={arXiv preprint arXiv:2505.08231},
  year={2025}
}

@article{su2023survey,
  title={A survey of maritime vision datasets},
  author={Su, Li and Chen, Yusheng and Song, Hao and Li, Wanyi},
  journal={Multimedia Tools and Applications},
  volume={82},
  number={19},
  pages={28873--28893},
  year={2023},
  publisher={Springer}
}

@article{jungbauer2025maritime,
  title={Maritime Vision Datasets for Autonomous Navigation: A Comparative Analysis},
  author={Jungbauer, Nico and Huang, Hai and Mayer, Helmut},
  journal={Maritime Technology and Research},
  volume={7},
  number={4},
  pages={Manuscript--Manuscript},
  year={2025}
}

@article{custom,
  title={An efficient model for small object detection in the maritime environment},
  author={Shao, Zeyuan and Yin, Yong and Lyu, Hongguang and Soares, C Guedes and Cheng, Tao and Jing, Qianfeng and Yang, Zhilin},
  journal={Applied Ocean Research},
  volume={152},
  pages={104194},
  year={2024},
  publisher={Elsevier}
}

@article{custom2,
  title={Vision-based maritime object detection covering far and tiny obstacles},
  author={Yoneyama, Ryota and Dake, Yuichiro},
  journal={IFAC-PapersOnLine},
  volume={55},
  number={31},
  pages={210--215},
  year={2022},
  publisher={Elsevier}
}

@article{abo,
  title={Aboships—an inshore and offshore maritime vessel detection dataset with precise annotations},
  author={Iancu, Bogdan and Soloviev, Valentin and Zelioli, Luca and Lilius, Johan},
  journal={Remote Sensing},
  volume={13},
  number={5},
  pages={988},
  year={2021},
  publisher={MDPI}
}

@article{liu_esod_2024,
  title={ESOD: efficient small object detection on high-resolution images},
  author={Liu, Kai and Fu, Zhihang and Jin, Sheng and Chen, Ze and Zhou, Fan and Jiang, Rongxin and Chen, Yaowu and Ye, Jieping},
  journal={IEEE Transactions on Image Processing},
  year={2024},
  publisher={IEEE}
}

@inproceedings{liu_ssd_2016,
  title={Ssd: Single shot multibox detector},
  author={Liu, Wei and Anguelov, Dragomir and Erhan, Dumitru and Szegedy, Christian and Reed, Scott and Fu, Cheng-Yang and Berg, Alexander C},
  booktitle={European conference on computer vision},
  pages={21--37},
  year={2016},
  organization={Springer}
}

@inproceedings{carion_end--end_2020,
  title={End-to-end object detection with transformers},
  author={Carion, Nicolas and Massa, Francisco and Synnaeve, Gabriel and Usunier, Nicolas and Kirillov, Alexander and Zagoruyko, Sergey},
  booktitle={European conference on computer vision},
  pages={213--229},
  year={2020},
  organization={Springer}
}

@inproceedings{zong_detrs_2023,
  title={Detrs with collaborative hybrid assignments training},
  author={Zong, Zhuofan and Song, Guanglu and Liu, Yu},
  booktitle={Proceedings of the IEEE/CVF international conference on computer vision},
  pages={6748--6758},
  year={2023}
}

@inproceedings{roh_sparse_2022,
title={Sparse {DETR}: Efficient End-to-End Object Detection with Learnable Sparsity},
author={Byungseok Roh and JaeWoong Shin and Wuhyun Shin and Saehoon Kim},
booktitle={International Conference on Learning Representations},
year={2022},
}

@inproceedings{zheng_less_2023,
  title={Less is more: Focus attention for efficient detr},
  author={Zheng, Dehua and Dong, Wenhui and Hu, Hailin and Chen, Xinghao and Wang, Yunhe},
  booktitle={Proceedings of the IEEE/CVF international conference on computer vision},
  pages={6674--6683},
  year={2023}
}

@inproceedings{simple_fpn,
  title={Exploring plain vision transformer backbones for object detection},
  author={Li, Yanghao and Mao, Hanzi and Girshick, Ross and He, Kaiming},
  booktitle={European conference on computer vision},
  pages={280--296},
  year={2022},
  organization={Springer}
}

@misc{roboflow,
  author= {Roboflow},
  title= {Roboflow: Organize, Label, and Deploy Computer Vision Datasets},
  howpublished = {\url{https://roboflow.com}},
  note= {Accessed: 2025-07-03}
}

\end{document}